\documentclass[a4paper,conference]{IEEEtran}

\usepackage{soul}
\usepackage{url}
\usepackage[hidelinks]{hyperref}
\usepackage[utf8]{inputenc}
\usepackage{graphicx}
\usepackage{amsmath}
\usepackage{booktabs}
\usepackage{algorithm}
\usepackage{algorithmic}
\usepackage{epsfig}
\usepackage{amssymb}
\usepackage{amsthm}
\usepackage{color}
\usepackage{dblfloatfix}

\newcommand{\figref}[1]{Fig.~\ref{fig:#1}}
\newcommand{\tabref}[1]{Table~\ref{tab:#1}}
\newcommand{\secref}[1]{Section~\ref{sec:#1}}

\newcommand{\ra}[1]{\renewcommand{\arraystretch}{#1}} 

\newcommand{\figlandmarksamples}{
\begin{figure}[t]
    \centering
    \includegraphics[width=\linewidth]{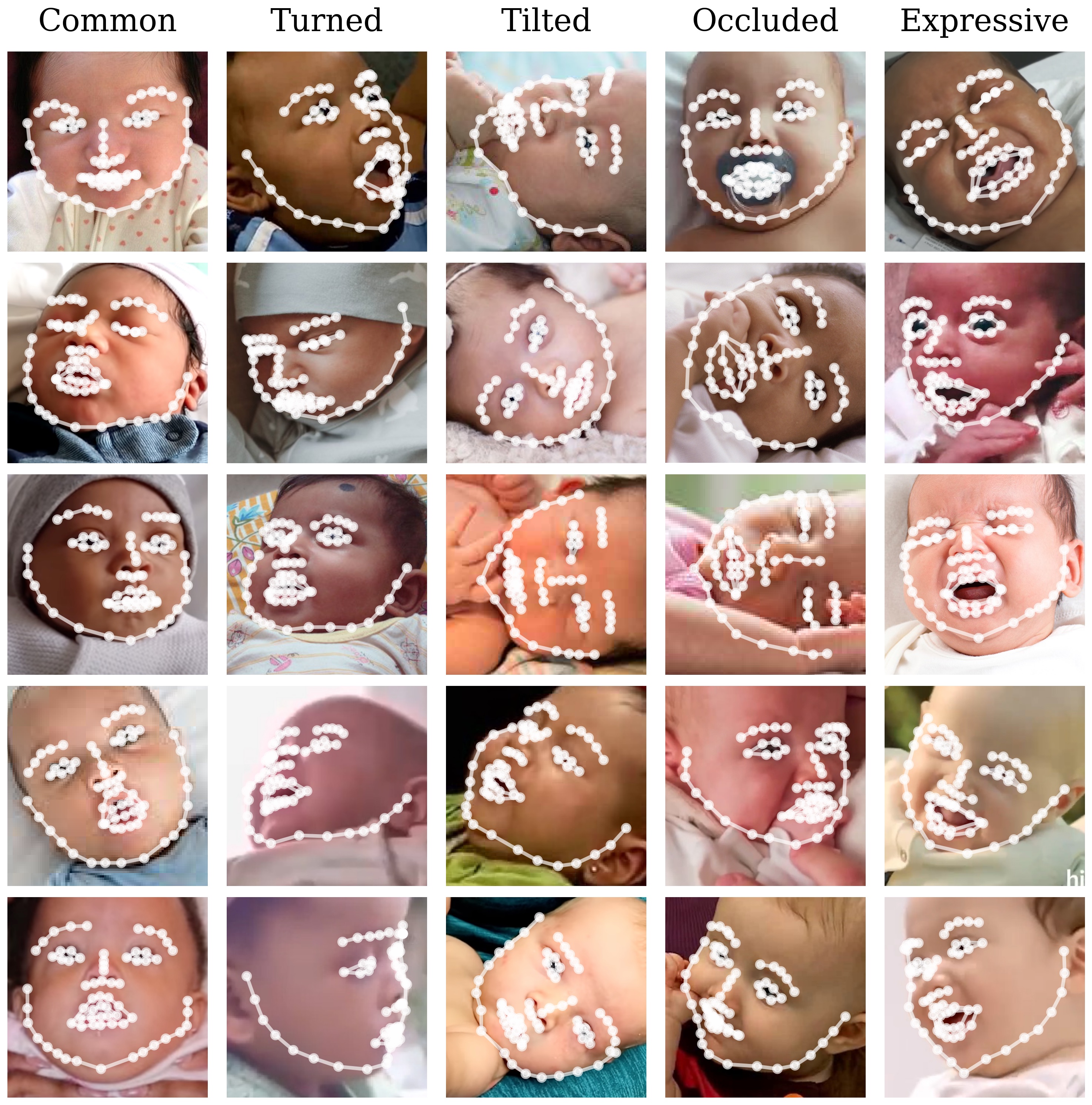}
    \vspace{-.25in}
    \caption{InfAnFace images and ground truth facial landmarks, grouped by annotated attributes.}
    \label{fig:landmark-samples}
    \vspace{-.2in}
\end{figure}
}

\newcommand{\figlandmarkpredictions}{
\begin{figure*}
    \centering
    \includegraphics[width=\linewidth]{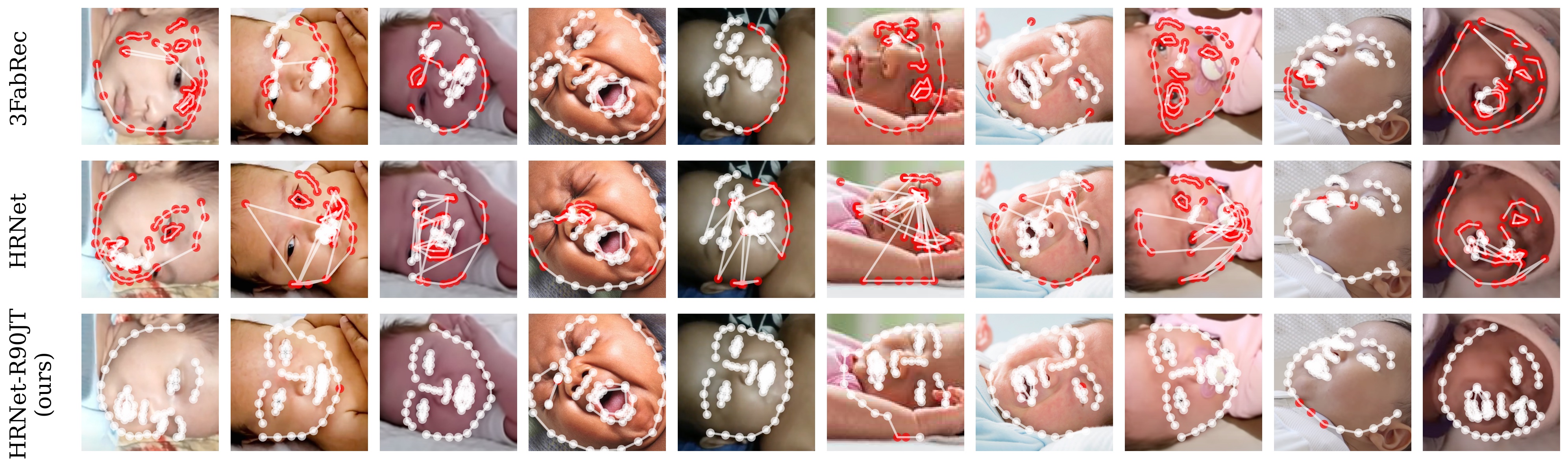}
    \vspace{-.2in}
    \caption{Model predictions on InfAnFace Test images, from 3FabRec, HRNet, and our state-of-the-art HRNet-R90JT model. Landmark predictions with high error are highlighted in red.}
    \label{fig:landmark-predictions}
    \vspace{-.2in}
\end{figure*}
}

\newcommand{\figcedcurves}{
\begin{figure}
    \centering
    \includegraphics[width=\linewidth]{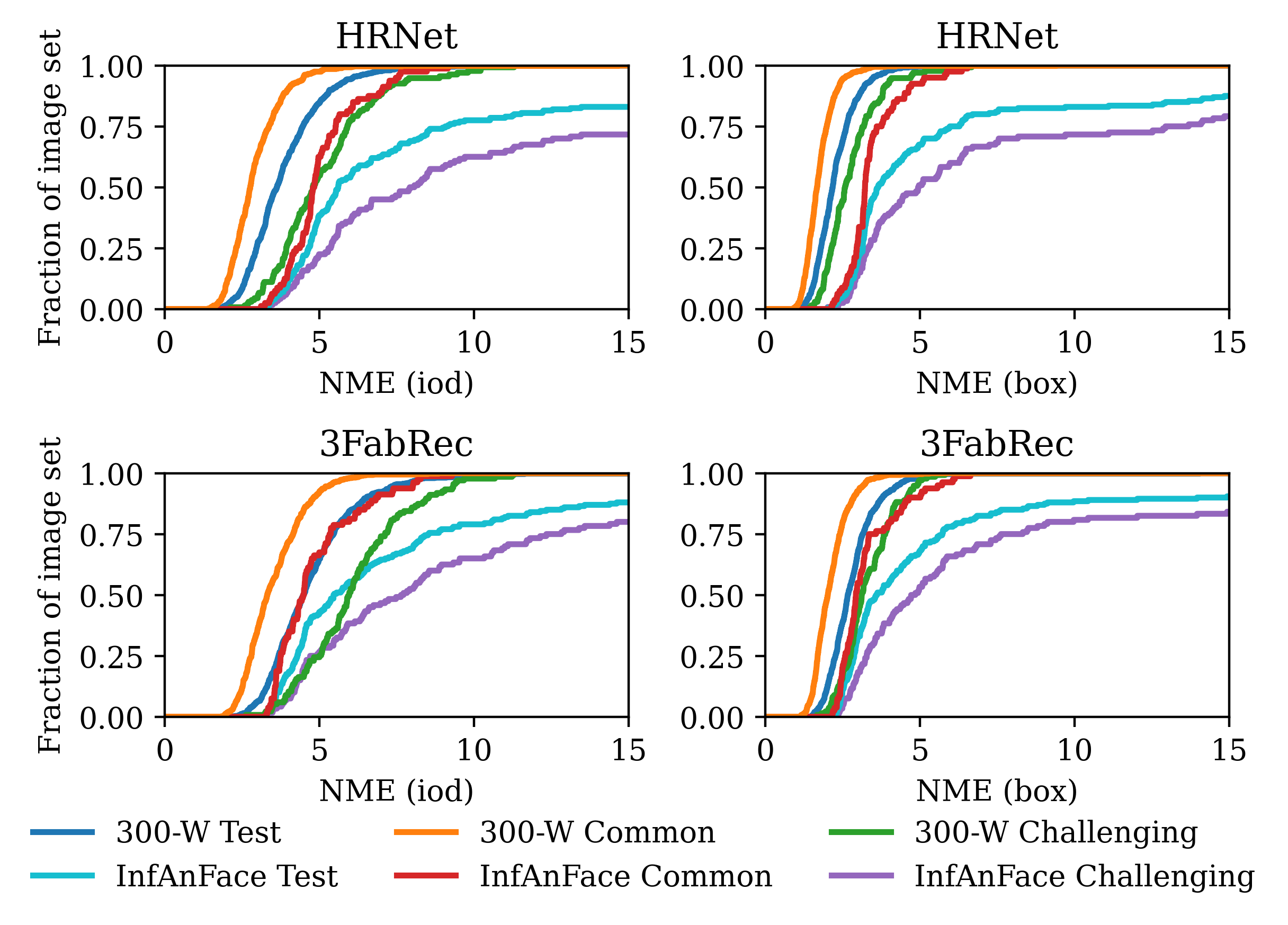}
    \vspace{-.25in}
    \caption{Cumulative normalized mean error (NME) curves for HRNet and 3FabRec facial landmark estimation models, highlighting the performance gap between infants (InfAnFace) and adults (300-W). Errors under both the interocular (iod) and minimal bounding box (box) normalization factors shown.}
    \vspace{-.25in}
    \label{fig:cumulative-distribution-curves}
\end{figure}
}

\newcommand{\figtsne}{
\begin{figure}
    \centering
    \includegraphics[width=\linewidth]{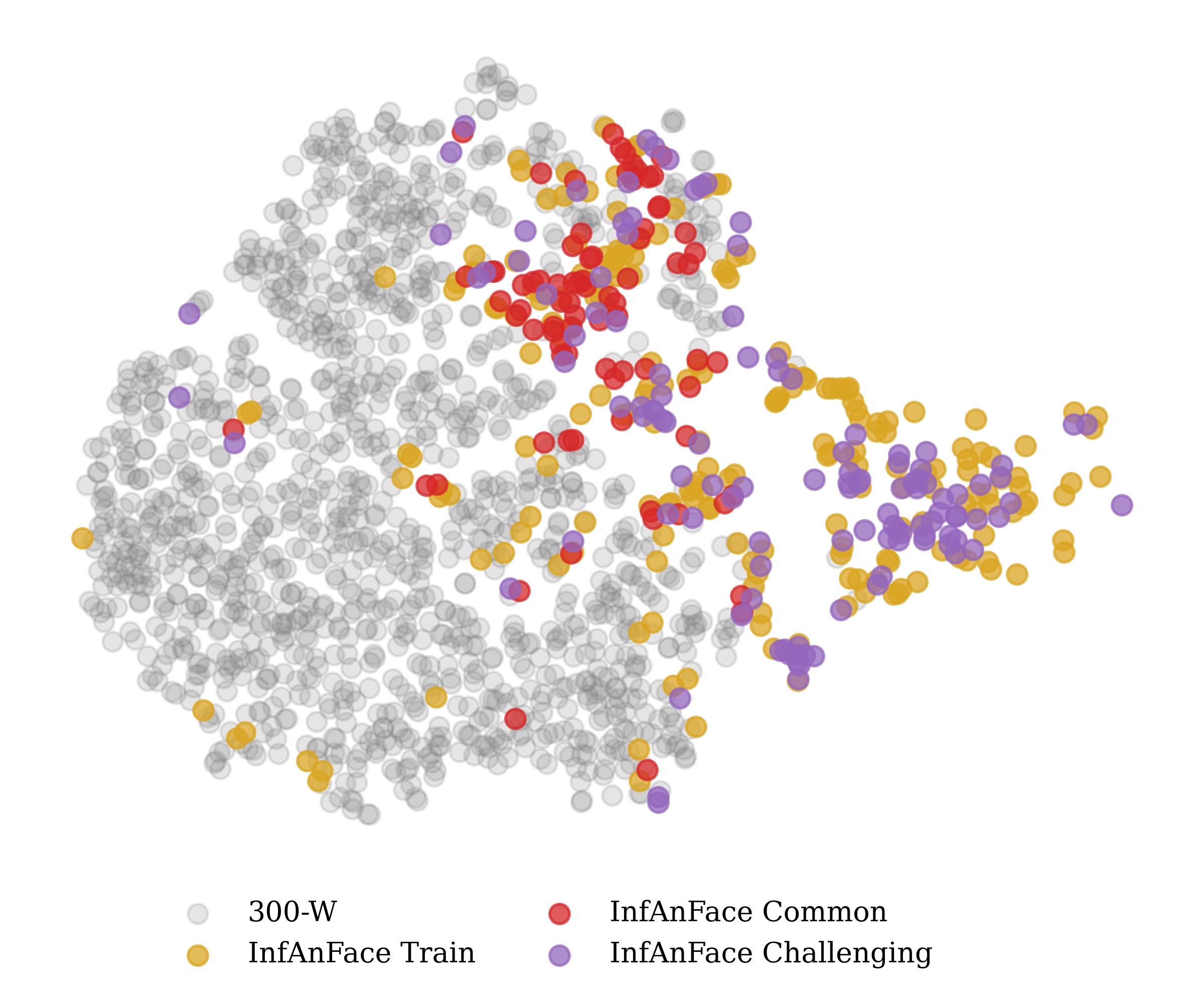}
     \vspace{-.25in}
    \caption{A t-SNE visualization of the internal HRNet representations of adult 300-W and InfAnFace images, highlighting the domain gap between them. InfAnFace Common is more closely integrated with the adult images compared to the more divergent InfAnFace Challenging. Together they comprise InfAnFace Test, which has roughly the same distribution as InfAnFace Train.}
    \vspace{-.2in}
    \label{fig:t-sne}
\end{figure}
}

\newcommand{\fighrnet}{
\begin{figure}
    \centering
    \includegraphics[width=\linewidth]{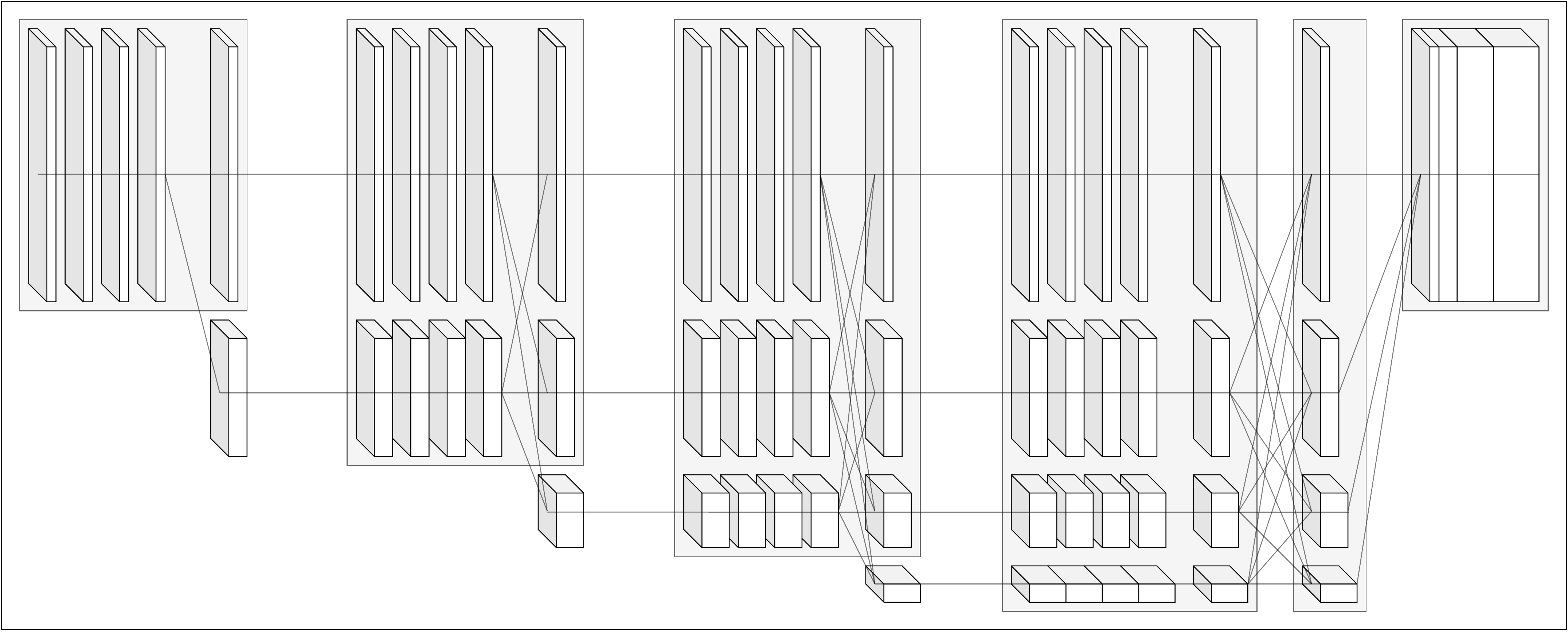}
     \vspace{-.25in}
    \caption{Structure of high-resolution net (HRNet) \cite{wang_deep_2021}, the backbone for our best models, featuring parallel multi-resolution convolutional layers.}
    \vspace{-.2in}
    \label{fig:hrnet}
\end{figure}
}

\newcommand{\figmeanfaceerrors}{
\begin{figure}[t]
    \centering
    \includegraphics[width=\linewidth]{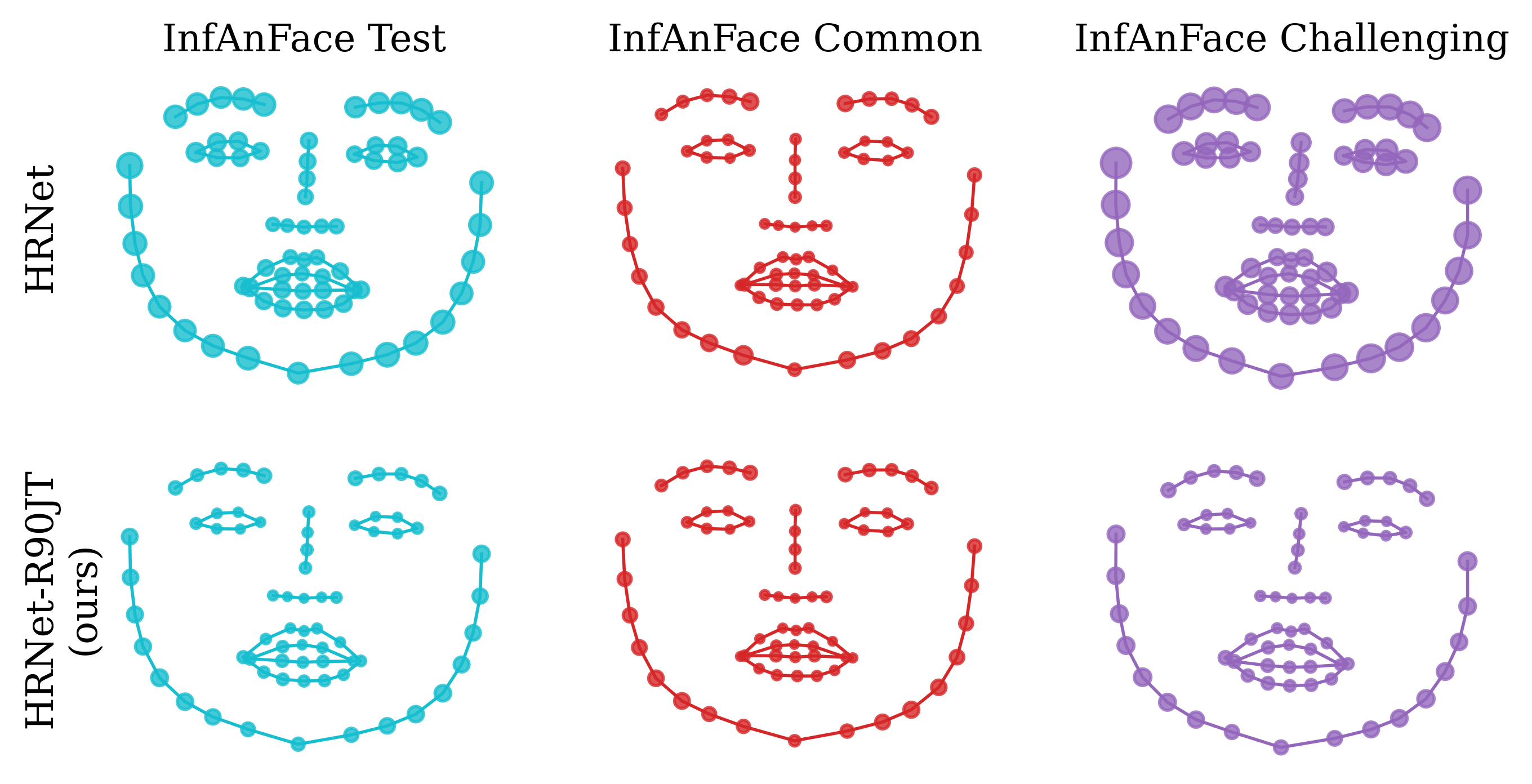}
    \vspace{-.3in}
    \caption{A visualization depicting the performance per landmark of HRNet against our HRNet-R90JT across various InfAnFace subset, with each landmark radius drawn in proportion to the subset-mean interocular-normalized error for that landmark. (Landmark positions are as in Supp. \figref{mean-faces}.)}
    \vspace{-.2in}
    \label{fig:mean-face-errors}
\end{figure}
}

\newcommand{\figmeanfaces}{
\begin{figure}[b]
    \centering
    \includegraphics[width=\linewidth]{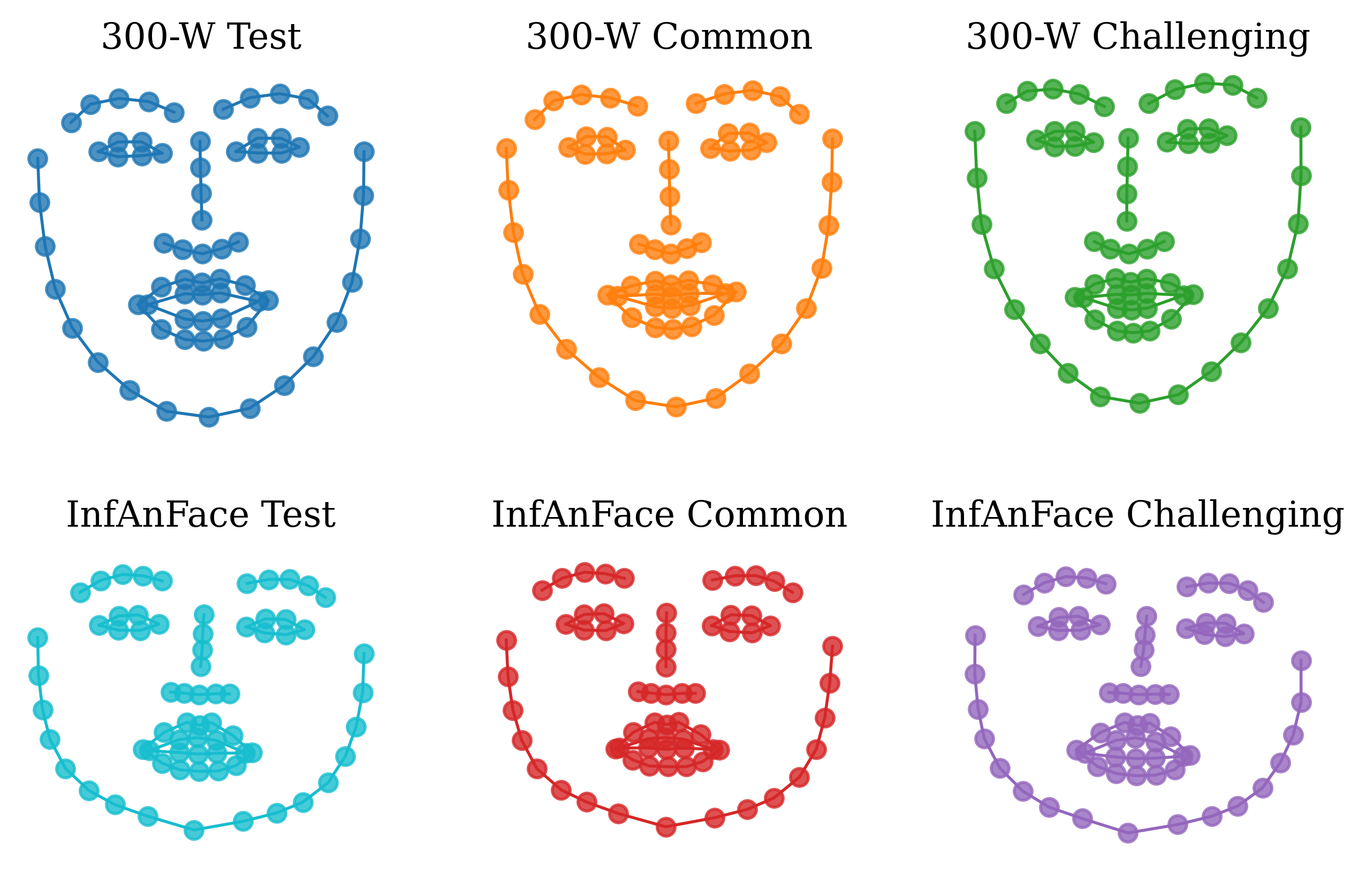}
    \vspace{-.3in}
    \caption{Mean ground truth landmarks across various adult 300-W and InfAnFace image sets, scaled to a common width. Infant faces appear squatter.}
    \label{fig:mean-faces}
\end{figure}
}

\newcommand{\figcedcurvesinfants}{
\begin{figure}[t]
    \centering
    \includegraphics[width=\linewidth]{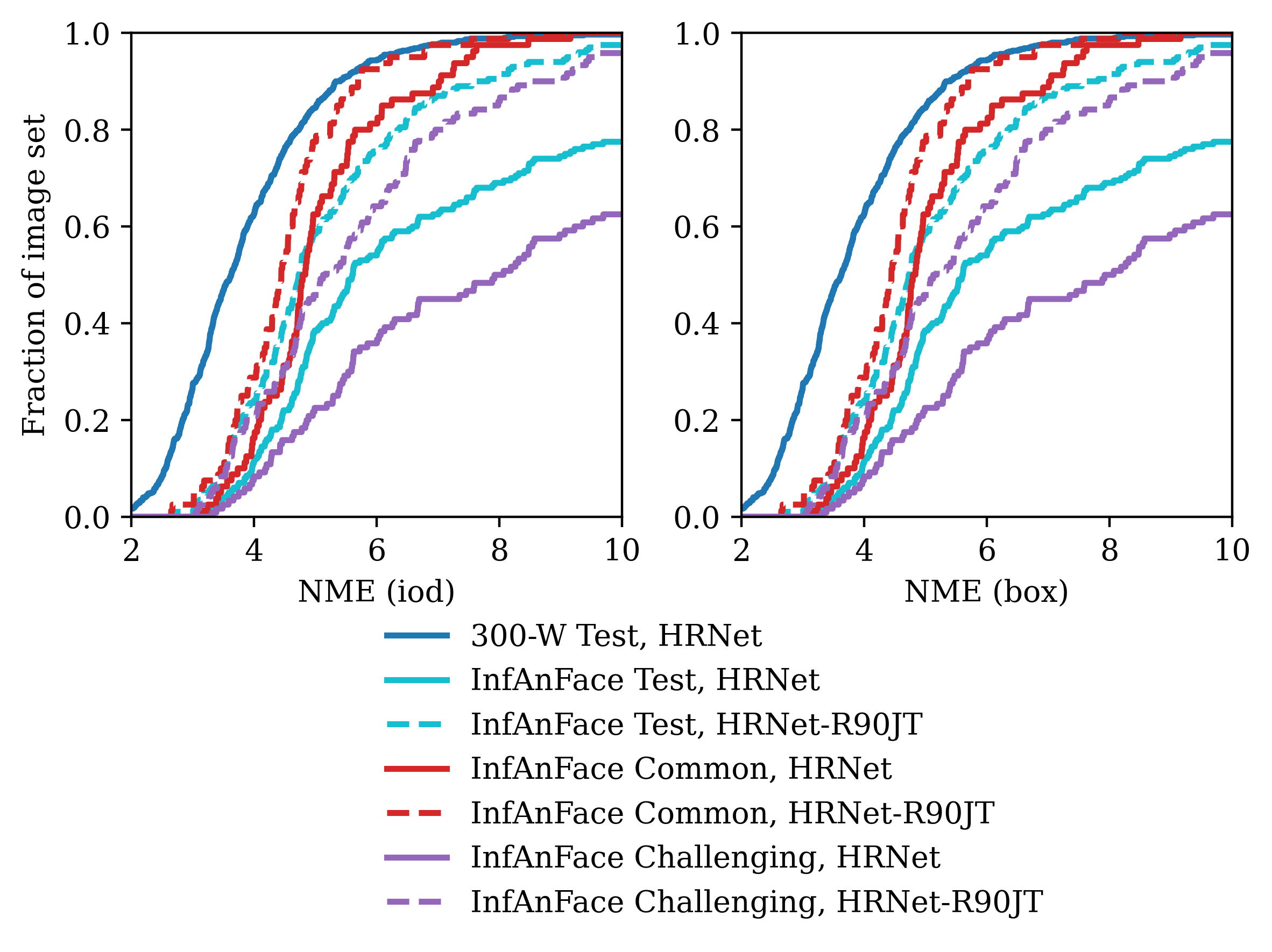}
    \vspace{-.3in}
    \caption{Cumulative normalized mean error (NME) curves showing the significant improvement of our HRNet-R90JT model (dashed lines) over HRNet (solid lines), across various InfAnFace Test subsets. The performance gains are especially notable on InfAnFace Challenging set. Performance of HRNet on an adult 300-W Test set is included for comparison. }
    \label{fig:cumulative-distribution-curves-infants}
\end{figure}
}
\usepackage{amssymb}
\usepackage{multicol}
\usepackage{multirow}
\usepackage{threeparttable}
\usepackage{bigdelim}

\newcommand{\tblattributes}{
\begin{table}[b]
    \footnotesize
    \centering
    \vspace{-0.2in}
    \caption{Attribute incidence rates in InfAnFace subsets}   
    \vspace{-0.1in}
    \begin{tabular}{@{}lrrr@{}}
        \toprule
        Attribute & \% of Inf. Train & \% of Inf. Test & \% of InfAnFace\\
        \midrule
        Turned & 28.6 & 12.5 & 20.2 \\ 
        Tilted & 28.6 & 36.5 & 32.0 \\
        Occluded & 26.2 & 14.0 & 20.2 \\
        Expressive & 37.1 & 14.5 & 26.1 \\ 
        \bottomrule
    \end{tabular}
    \label{tab:attributes}
\end{table}
}

\newcommand{\tblfaceratios}{
\begin{table}[b]
    \footnotesize
    \centering
    \caption{Geometric values across adult 300-W and InfAnFace images} 
    \vspace{-0.1in}
    \begin{tabular}{@{}lcc@{}}
        \toprule
        Image Set & $\frac{\mu(\text{min. box width})}{\mu(\text{min. box height})}$ & $\mu\left(\frac{\text{min. box size}}{\text{interocular dist.}}\right)$ \\
        \midrule
        300-W Test & 1.01 & 1.69 \\
        300-W Common & 1.03 & 1.67 \\
        300-W Challenging & 1.02 & 1.85 \\
        \midrule
        InfAnFace Test & 1.28 & 1.53 \\
        InfAnFace Common & 1.28 & 1.48 \\
        InfAnFace Challenging & 1.27 & 1.56 \\
        \bottomrule
    \end{tabular}
    \label{tab:face-ratios}
\end{table}
}

\newcommand{\tblnme}{
\begin{table}
  \caption{Landmark estimation normalized mean error ({\upshape NME}, $\downarrow$ is better) under {\upshape iod} and {\upshape box} normalizations, on adult 300-W and InfAnFace images}
    \centering
    \ra{1.3}
    \resizebox{\columnwidth}{!}{\begin{tabular}{@{}lrrcrrcrr@{}}
        \toprule
        & \multicolumn{2}{c}{\text{300-W Test}} & \phantom{a} & \multicolumn{2}{c}{\text{300-W Comm.}} & \phantom{a} & \multicolumn{2}{c}{\text{300-W Chal.}}\\ \cmidrule{2-3} \cmidrule{5-6} \cmidrule{8-9}
        Model & iod $\downarrow$ & box $\downarrow$ && iod $\downarrow$ & box $\downarrow$ && iod $\downarrow$ & box $\downarrow$\\
        \midrule
        3FabRec (\textit{CVPR} '20) & 3.85 & 2.24 && 3.40 & 2.04 && 5.74 & 3.10\\
        HRNet (\textit{TPAMI} '21) & 3.85 & 2.28 && 2.92 & 1.76 && 5.13 & 2.77\\
        \midrule
        & \multicolumn{2}{c}{\text{InfAnFace Test}} & \phantom{a} & \multicolumn{2}{c}{\text{InfAnFace Comm.}} & \phantom{a} & \multicolumn{2}{c}{\text{InfAnFace Chal.}}\\ \cmidrule{2-3} \cmidrule{5-6} \cmidrule{8-9}
        Model & iod $\downarrow$ & box $\downarrow$ && iod $\downarrow$ & box $\downarrow$ && iod $\downarrow$ & box $\downarrow$\\
        \midrule
        3FabRec (\textit{CVPR} '20) & 12.59 & 8.13 && \textit{4.93} & \textit{3.36} && 17.69 & 11.31\\
        3FabRec-JT & 9.28 & 6.04 && \textbf{4.90} & \textbf{3.34} && 12.20 & 7.84\\
        3FabRec-FT& 9.00 & 5.86 && 5.30 & 3.59 && 11.47 & 7.37\\
        \textit{3FabRec-R90} & \textit{8.45} & \textit{5.52} && 5.45 & 3.71 && \textit{10.45} & \textit{6.73}\\
        \textbf{3FabRec-R90JT} & \textbf{8.00} & \textbf{5.22} && 5.12 & 3.49 && \textbf{9.93} & \textbf{6.39}\\
        \midrule
        HRNet (\textit{TPAMI} '21) & 12.82 & 8.35 && 5.07 & 3.45 && 17.98 & 11.62\\
        HRNet-JT & 5.95 & 3.87 && \textbf{4.47} & \textbf{3.03} && 6.94 & 4.43\\
        HRNet-FT& 5.90 & 3.85 && \textit{4.48} & \textit{3.04} && 6.84 & 4.38\\ 
        HRNet-R90 & 5.90 & 3.87 && 4.88 & 3.31 && 6.58 & 4.24\\
        \textbf{HRNet-R90JT} (ours) & \textbf{5.30} & \textbf{3.47} && 4.52 & 3.07 && \textbf{5.82} & \textbf{3.73}\\
        \textit{HRNet-R90FT} (ours) & \emph{5.40} & \emph{3.52} && 4.68 & 3.17 && \emph{5.88} & \emph{3.78}\\
        \bottomrule
    \end{tabular}}
    \vspace{-0.2in}
    \label{tab:nme-table}
\end{table}
}

\newcommand{\tblfrauc}{
\begin{table}
  \caption{Landmark estimation failure rate ({\upshape FR}, out of 100, $\downarrow$ is better) and area under the curve ({\upshape AUC}, out of 100, $\uparrow$ is better) at {\upshape $\text{NME}_\text{iod}=10$}, on adult 300-W and InfAnFace images}
    \centering
    \ra{1.3}
    \resizebox{\columnwidth}{!}{\begin{tabular}{@{}lrrcrrcrr@{}}
        \toprule
        & \multicolumn{2}{c}{\text{300-W Test}} & \phantom{a} & \multicolumn{2}{c}{\text{300-W Comm.}} & \phantom{a} & \multicolumn{2}{c}{\text{300-W Chal.}}\\ \cmidrule{2-3} \cmidrule{5-6} \cmidrule{8-9}
        Model & FR $\downarrow$ & AUC $\uparrow$ && FR $\downarrow$ & AUC $\uparrow$ && FR $\downarrow$ & AUC $\uparrow$\\ 
        \midrule
        3FabRec (\textit{CVPR} '20) & 1.17 & 52.72 && 0.00 & 64.73 && 2.22 & 38.95 \\
        HRNet (\textit{TPAMI} '21) & 0.33 & 61.54 && 0.18 & 70.83 && 2.22 & 48.84 \\
        \midrule
        & \multicolumn{2}{c}{\text{InfAnFace Test}} & \phantom{a} & \multicolumn{2}{c}{\text{InfAnFace Comm.}} & \phantom{a} & \multicolumn{2}{c}{\text{InfAnFace Chal.}}\\ \cmidrule{2-3} \cmidrule{5-6} \cmidrule{8-9}
        Model & FR $\downarrow$ & AUC $\uparrow$ && FR $\downarrow$ & AUC $\uparrow$ && FR $\downarrow$ & AUC $\uparrow$\\ 
        \midrule
        3FabRec (\textit{CVPR} '20) & 21.00 & \textit{36.66} && \textit{0.00} & \textbf{51.69} && 35.00 & 26.63 \\
        3FabRec-JT & 17.00 & \textbf{37.34} && 1.25 & \textit{51.38} && 27.50 & \textbf{27.98}\\
        3FabRec-FT & \textit{14.50} & 35.29 && \textit{0.00} & 46.88 && \textit{24.17} & \textit{27.57}\\
        3FabRec-R90 & 18.00 & 33.21 && 1.25 & 46.15 && 29.17 & 24.58\\
        3FabRec-R90JT & \textbf{14.00} & 36.16 && 1.25 & 49.43 && \textbf{22.50} & 27.32\\
        \midrule
        HRNet (\textit{TPAMI} '21) & 22.50 & 34.75 && 0.00 & 49.30 && 37.50 & 25.06\\
        HRNet-JT & 7.00 & 45.52 && 0.00 & \textbf{55.32} && 11.67 & 38.99\\
        HRNet-FT & 6.00 & 45.91 && 0.00 & \textit{55.22} && 10.00 & 39.71 \\
        HRNet-R90 & 5.00 & 43.75 && 0.00 & 51.18 && 8.33 & 38.80\\
        \textbf{HRNet-R90JT} (ours) & \textbf{2.50} & \textbf{48.07} && 0.00 & 54.69 && \textbf{4.17} & \textbf{43.65} \\
        \textit{HRNet-R90FT} (ours) & \emph{3.00} & \emph{46.87} && 0.00 & 53.23 && \emph{5.00} &\emph{42.63}\\
        \bottomrule
    \end{tabular}}
    \vspace{-0.2in}
    \label{tab:fr-auc-table}
\end{table}
}

\newcommand{\supp}{
\clearpage
\newcommand{\beginsupplement}{%
        \setcounter{table}{0}
        \setcounter{equation}{0}
        \renewcommand{\theequation}{S\arabic{equation}}
        \setcounter{section}{0}
        \renewcommand{\thetable}{S\arabic{table}}%
        \setcounter{figure}{0}
        \renewcommand{\thefigure}{S\arabic{figure}}%
}

\newcommand{\independent}{\protect\mathpalette{\protect\independenT}{\perp}}
\def\independenT##1##2{\mathrel{\rlap{$##1##2$}\mkern2mu{##1##2}}}
\renewcommand\thesection{\Alph{section}}
\beginsupplement
\newcommand{\MYhref}[3][blue]{\href{##2}{\color{##1}{##3}}}

\section{Supplementary Materials}
\label{sec:supplementary}

\subsection{Mean adult and infant faces}
\label{sec:mean-faces}

A broad view of differences between adult and infant facial landmark geometries is provided by \figref{mean-faces}, which shows the mean positions of ground truth landmarks within various 300-W and InfAnFace image sets. Mean positions reflect both intrinsic facial geometry \emph{and} non-intrinsic facial pose (such as head orientation), so conclusions must be drawn cautiously. Still, adult faces certainly \textit{appear} taller, with eyes set higher up the nose and jawlines more oblong. This corroborates our observations from \tabref{face-ratios} in \secref{geometric-comparisons}.
\figmeanfaces

\subsection{Cumulative error distribution curves for infant landmark estimation models}
\label{sec:ced-curves}

\figref{cumulative-distribution-curves-infants} displays the cumulative error distribution curves for both HRNet and our best-performing new model, HRNet-R90JT. It shows performance gains across the board, but especially on the InfAnFace Challenging set, nearly eliminating the failure rate at $\text{NME}_\text{iod}=10$. The overall performance of HRNet-R90JT on InfAnFace Common even approaches the performance of HRNet on 300-W Test, a standard adult image set.
\figcedcurvesinfants

\subsection{Definition of average precision for facial detection}
\label{sec:average-precision}

We elaborate on the definition of the \textit{average precision} metric for facial detection from \secref{face-detection}. The output of a facial detection algorithm consists of bounding boxes together with probabilistic confidence scores attached. For each $0\leq\alpha<1$, predictions with confidence scores greater than $\alpha$ are considered to be true positives if they overlap with a ground truth box with an area at least half that of their union, and as a result a precision and recall score can be computed for that $\alpha$. By varying $\alpha$, a precision-recall curve is generated, and the average precision is defined to be the interpolated area under that curve. See \cite{Everingham10} for details.

}

\begin{document}

%
\title{InfAnFace: Bridging the Infant--Adult Domain Gap \\ in Facial Landmark Estimation in the Wild}

\author{
\IEEEauthorblockN{Michael Wan\IEEEauthorrefmark{1}\IEEEauthorrefmark{2}, Shaotong Zhu\IEEEauthorrefmark{2}, Lingfei Luan\IEEEauthorrefmark{2}, Gulati Prateek\IEEEauthorrefmark{2}, Xiaofei Huang\IEEEauthorrefmark{2}, \\Rebecca Schwartz-Mette\IEEEauthorrefmark{3}, Marie Hayes\IEEEauthorrefmark{3}, Emily Zimmerman\IEEEauthorrefmark{2}, Sarah Ostadabbas\IEEEauthorrefmark{2}\IEEEauthorrefmark{4}}
\IEEEauthorblockA{\IEEEauthorrefmark{1}Roux Institute, Portland, ME, USA}
\IEEEauthorblockA{\IEEEauthorrefmark{2}Northeastern University, Boston, MA, USA}
\IEEEauthorblockA{\IEEEauthorrefmark{3}University of Maine, Orono, ME, USA}
\IEEEauthorblockA{\IEEEauthorrefmark{4}Corresponding author: \texttt{ostadabbas@ece.neu.edu}}
}

\maketitle

\begin{abstract}
We lay the groundwork for research in the algorithmic comprehension of infant faces, in anticipation of applications from healthcare to psychology, especially in the early prediction of developmental disorders. Specifically, we introduce the first-ever dataset of infant faces annotated with facial landmark coordinates and pose attributes, demonstrate the inadequacies of existing facial landmark estimation algorithms in the infant domain, and train new state-of-the-art models that significantly improve upon those algorithms using domain adaptation techniques. We touch on the closely related task of facial detection for infants, and also on a challenging case study of infrared baby monitor images gathered by our lab as part of in-field research into the aforementioned developmental issues\footnote{The dataset and model code is available at:  \href{https://github.com/ostadabbas/Infant-Facial-Landmark-Detection-and-Tracking}{https://github.com/ostadabbas/Infant-Facial-Landmark-Detection-and-Tracking}.}.
\end{abstract}


\section{Introduction}
\label{sec:intro}

The development of effective facial landmark estimation algorithms for infants will open up new avenues of research in healthcare and cognitive psychology. For instance, links between early infant motor or oral function and subsequent developmental disruptions---such as autism spectrum disorder \cite{ali2020early}, cerebral palsy \cite{centers2012data}, and pediatric feeding disorders \cite{lindberg1991early}---inspire a tantalizing vision of algorithmic screening or even discovery of prodromal (or pre-symptomatic) risk markers, grounded on automatic facial landmark estimation. However, while vision-based methods have been wildly successful on adult faces, hardly any analogous models exist in the domain of infant faces. Part of the problem is that infants rarely appear in human images ``in the wild,'' and when they do, their faces are often small and obscured and hence not useful for the purposes of facial landmark estimation. 

In this paper, we present multi-pronged efforts intended to initiate and foster an ecosystem of research in infant facial landmark estimation. These efforts include:
\begin{itemize}
    \item  the creation of the first-ever \textbf{Inf}ant \textbf{An}notated \textbf{Face}s (InfAnFace) dataset, consisting of 410 images of infant faces with labels for 68 facial landmark locations and various pose attributes (see \figref{landmark-samples}); 
    \item  the organization of InfAnFace into a Train set, intended for machine learning, and an independently-sourced Test set, intended to serve as the field benchmark; 
    \item  the demonstration of an infant--adult domain gap for existing facial landmark estimation algorithms; 
    \item successful steps towards bridging this gap using domain adaptation techniques under ``small-data'' conditions, achieving state-of-the-art estimation performance; 
    \item  a brief discussion on algorithmic face detection for infants, a prerequisite task for landmark estimation; 
    and
    \item  a report on a challenging real-life application of our algorithms to infrared baby monitor footage captured in-field by our lab. 
\end{itemize}
Our goal is to cast a wide net and encourage further research along each of these lines of inquiry. 

\figlandmarksamples

\section{Related work}
\label{sec:related}

Our work relates to a broad swath of prior computer vision research on faces. There are several datasets of faces ``in the wild,'' used to   train and benchmark facial landmark estimation algorithms. Datasets featuring 68 facial landmarks adhering to the Multi-PIE layout \cite{gross_multi-pie_2010}, which we adopt, include the Helen dataset of 2000 training and 300 test images, the annotated faces in the wild (AFW) dataset of 205 images, featuring large variation in head poses and multiple faces per image \cite{zhu2012face}, and various well-known datasets stemming from the 300 Faces In-the-Wild Challenge (300-W) \cite{sagonas_300_2016}, which we make extensive use of below. Infant faces are hardly represented at all, for instance comprising of only 1.4\% of the diverse AFW dataset.

Facial landmark estimation has a long history, with many pioneering models based on fitting face meshes or face models recently surpassed by pure convolutional neural network (CNN) regression methods (see surveys in  \cite{jin_face_2017} and \cite{wu_facial_2019}). While in principle, face model methods could help with the wide poses inherent in the infant domain, we are only aware of one attempt to build such a model for infants from 3D scans \cite{morales_spectral_2020}, and it is not available for public use. For our study, we stick with flexible and more powerful CNN regression models. We test and modify the well-established high-resolution network (HRNet) model, which uses multi-resolution CNN blocks to carry high-fidelity representations throughout the entire network \cite{wang_deep_2021}. We also test with the recent 3FabRec model \cite{browatzki_3fabrec_2020}, which achieves few-shot facial landmark localization via unsupervised autoencoder pre-training. 

Face detection, a prerequisite for facial landmark estimation, is another mature field, again dominated by CNN methods (see \cite{minaee_going_2021} for a survey). We perform tests on infant faces using the recent state-of-the-art RetinaFace model \cite{deng_retinaface_2019}. Finally, recent work by our team tackles the closely-related problem of infant \textit{body pose} estimation, using adversarial domain adaptation methods \cite{huang2021infant}.

\section{The InfAnFace Dataset}
\label{sec:dataset}

To establish the first dataset of \textbf{In}fant \textbf{An}notated \textbf{Face}s (InfAnFace), a total of 410 images were sourced and annotated with 68 facial landmarks plus pose attributes, by a team of researchers, to encourage variety and independence. Infants are only featured in a small fraction of human photographs, and when they do appear, their faces are often small or obscured, so gathering a sizable set of images useful for facial landmark estimation algorithms with deep structures is non-trivial. Facial annotations are also more difficult for infants because features are less well-defined, poses are more varied, and natural obstructions like pacifiers are more prevalent. Thus, we believe these diverse sets of well-annotated images represent a hard-earned contribution to a field of study which is itself in its infancy. The results reported later in this paper demonstrate the effectiveness of InfAnFace for machine learning training and testing.

\textit{Landmark annotations.} The 68 facial landmark annotations adhere to the industry standard Multi-PIE layout \cite{gross_multi-pie_2010}. We have also included coordinates of the \textit{minimal bounding box}, the smallest upright box containing all of the landmarks. Following industry conventions for 2D-based annotations, landmarks obscured by the face itself (e.g. when turned) were assigned to the nearest point on the face which \emph{is} visible in the image, as the true projected position is hard to estimate. We employed both a standard annotation tool \cite{dutta2019vgg}, and the specialized human--artificial intelligence hybrid tool AH-CoLT \cite{huang2019ah}, with the landmark predictions of the FAN model \cite{bulat_how_2017} serving as the starting point for the human annotations.

\textit{Face attributes.} To facilitate analysis, we included binary annotations for each image, indicating whether infant's face is: \emph{turned} (if the eyes, nose, and mouth are not clearly visible), \emph{tilted} (if the head axis, projected on the image plane, is $45^{\circ}$ or more beyond upright), \emph{occluded} (if landmarks are covered by body parts or objects), and excessively \emph{expressive} (if the facial muscles are tense, as when crying, laughing, etc.).

\textit{Sourcing and the Test--Train split.} InfAnFace images were sourced from Google Images and YouTube via a wide range of search queries to obtain a diversity of appearances, poses, expressions, scene conditions, and image quality. Approximately two-fifths of the infants represented appear to be non-white. For our machine learning experiments, we split InfAnFace into dedicated training and test sets, with the following composition: \textbf{InfAnFace Train}, consisting of 210 images, with 51- and 55-image batches drawn respectively from Google and YouTube, and a 105-image batch drawn from a specialized search on YouTube for infant formula advertisements; and \textbf{InfAnFace Test}, consisting of 200 images, with two 100-image batches drawn respectively from Google and YouTube. These five ``batches'' were sourced and annotated by rotating configurations of researchers, ensuring a level of variety and independence between them and hence between InfAnFace Train and InfAnFace Test as well.

\tblattributes
\textit{Attributes and the Common--Challenging split.} The breakdown of annotated attributes across InfAnFace and its Train and Test subsets, shown in \tabref{attributes}, demonstrates that all three sets exhibit a healthy diversity of ideal and adverse conditions. Each attribute is present in each set to a large enough degree to be useful for training or analysis, but the distributional differences between InfAnFace Train and Test also highlight their independence. To aid interpretability of experiments performed on InfAnFace Test, we partition it further into two subsets based on the annotated attributes:  \textbf{InfAnFace Common}, the subset 80 images with faces free from all adverse attributes; and \textbf{InfAnFace Challenging}, the remaining set of 120 images exhibiting at least one of them. 

\textit{To sum up, InfAnFace (410 images) is split into InfAnFace Train (210) and InfAnFace Test (200), and InfAnFace Test into InfAnFace Common (80) and InfAnFace Challenging (120).}

\section{Illustrating The Infant-Adult Domain Gap}
\label{sec:infant-adult-gap}

We examine the facial landmark domain gap between infants and adults by comparing InfAnFace Test with three predominantly adult image sets: 300-W Test (600 images), 300-W Common (554 images), and 300-W Challenging (135 images) \cite{sagonas_300_2016}. Our definitions for InfAnFace Test, Common, and Challenging loosely track with these 300-W sets, with Common featuring relatively easy poses, Test a balanced mix of poses, and Challenging the most difficult poses. The 300-W images are partially taken from earlier datasets so a range of sources is represented. We start with a brief comparison of geometric attributes, and then turn to a study of facial landmark estimation model performance on InfAnFace vs. 300-W, before capping off the section with a visualization of one of the model's internal representations of each image.

\subsection{Geometric observations}
\label{sec:geometric-comparisons}

\tblfaceratios

We record some geometric measures of interest across various adult and infant subsets in \tabref{face-ratios}. The first column shows that the aspect ratios of the minimal bounding boxes for mean adult faces are $\sim1:1$, compared to $\sim5:4$ for infants. The second column of \tabref{face-ratios} records the mean ratios of two commonly used normalization factors in facial landmark estimation. We will discuss these normalization factors and the significance of these values in the next section. See Supp. \secref{mean-faces} for a visual comparison of facial geometries.

\subsection{Benchmarking facial landmark estimation}
\label{sec:adult-model-performance}

\figlandmarkpredictions

In order the demonstrate the gap in facial landmark estimation between adults and infants, and to establish baseline performance metrics, we performed predictions using two recent 2D facial landmarking models, HRNet \cite{wang_deep_2021} (with the pretrained HRNetV2-W18 model) and 3FabRec \cite{browatzki_3fabrec_2020} (with the pretrained \texttt{lms\_300w} model). A selection of predicted landmarks is shown in \figref{landmark-predictions}, which also includes improved predictions from our own model (described in \secref{infant-landmarking-models}).

The main error metric for facial landmark estimation is the \textit{normalized mean error (NME)}: the mean Euclidean distance between each predicted landmark and the corresponding ground truth landmark, divided by a normalization factor with the same units (pixels), to achieve scale-independence. We consider two normalization factors: the \emph{interocular distance (iod)}, defined as the distance between the two outer corners of the eyes, and the \emph{minimal bounding box size (box)}, defined as the geometric mean of its height and width. The means of the ratios of these two normalization factors across various infant and adult sets is recorded in the second column of \tabref{face-ratios}. Those values show that the gap between metric performance on adults and infants will be greater on average under the box norm, compared to the interocular norm. Neither error metric is likely to be domain invariant, so using both provides a more rounded comparison. Note that although NME is sometimes reported as a percentage, it can exceed 100 in value.

\tblnme
\tblfrauc

We tabulate the mean NMEs under both normalizations and of both model's landmark predictions across various adult and infant image sets in \tabref{nme-table}. Further characterizations of landmark estimation performance include the \textit{failure rate (FR)} of images in the dataset with NME greater than a set threshold, and the \textit{area under the curve (AUC)} of the cumulative NME distribution up to a set threshold. \tabref{fr-auc-table} reports the FR and AUC for NME normalized by the interocular distance, with a threshold of $\text{NME} = 10$. (Both tables also include results from our improved models, to be discussed in \secref{infant-landmarking-models}.) \figref{cumulative-distribution-curves} plots the cumulative NME distribution curves themselves, under both normalizations, and with a threshold of $\text{NME} = 15$ for greater context.

\figcedcurves

These results show a significant performance gap between adult and infant domains as a whole, which in line with our expectations from \tabref{face-ratios} is more pronounced under the box norm. Within each domain, performance degrades from the Common to Test to Challenging sets. Furthermore, poor performance on InfAnFace Test seems largely attributable to the difficulty of the InfAnFace Challenging subset. These considerations, as well as a visual inspection of landmark predictions as in \figref{landmark-predictions}, expose adverse conditions defining the InfAnFace Challenging subset, such as tilt and occlusion, as likely causes of poor facial landmark estimations on infant faces. We believe, though, that such conditions are endemic to infant images captured in the wild, and thus, infant-focused algorithms should seek to overcome them. 

We note that while the performance results on InfAnFace Challenging are the most dramatic, the tables and graphs do also reveal a smaller but definite performance gap between results on InfAnFace Common and the adult image sets, particularly the 300-W Common and 300-W Test sets which arguably serve as fairer points of comparison. This suggests that even in more ideal pose conditions, a domain gap exists.

\subsection{t-SNE visualization}

\figtsne

An elegant visual companion to the preceding analysis can be found in the t-distributed stochastic neighbor embedding (t-SNE) plots in \figref{t-sne}, which offer a glimpse into the how the HRNet neural network ``perceives'' each image relative to one another. Each infant or adult image is processed by HRNet into a $270\times64\times64$-dimensional vector (before the regression head), and the set of these representations is compressed by the t-SNE algorithm \cite{van_der_maaten_visualizing_2008} into a set of two-dimensional coordinates for each image, with relative relationships probabilistically preserved. This set of coordinates is plotted in \figref{t-sne}, with image set membership highlighted by color\footnote{We employed the scikit-learn \cite{scikit-learn} implementation of t-SNE, with perplexity $50$ and otherwise default settings.}. In line with our earlier observations, the t-SNE distribution highlights a general gap between adult and infant image sets, and within InfAnFace Test, a split between the tamer InfAnFace Common subset and the more divergent InfAnFace Challenging subset. InfAnFace Test as a whole is distributed similarly to InfAnFace Train.

\section{Bridging The Infant-Adult Domain Gap}
\label{sec:infant-landmarking-models}

We now turn to our efforts to solve facial landmark estimation for infant faces. Since there exist strong models and ample data in the adult domain, for this inceptive paper we apply the domain adaptation tools of joint-training and fine-tuning, in addition to targeted data augmentation tools based on insights from our analysis of the domain gap. We have already exhibited the effectiveness of our resulting algorithm in \figref{landmark-predictions}, and our quantitative analysis will show that, for instance, our modified HRNet model performs with a failure rate of 2.50\% at $\text{NME}_\text{iod} = 10$ on InfAnFace Test, drastically improving upon the 22.50\% for the original HRNet model, and even approaching the 1.17\% failure rate of the original HRNet on the adult 300-W Test set. Note that commonly used adversarial discriminative domain adaptation methods like \cite{ganin_domain-adversarial_2016} or \cite{tzeng_adversarial_2017} work well for classification but are typically ineffective in the regression domain adaptation setting, where the covariate shift assumption usually holds (as discussed in \cite{kivinen_domain_2011} and especially \cite{de_mathelin_adversarial_2021}). So we believe that our methods yield results competitive with what can be obtained from general domain adaptation tools applicable in the landmark estimation setting, and serve as strong baselines for the nascent field.

\subsection{Training implementations}

\fighrnet

We started with two powerful state-of-the-art models, the high-resolution net (HRNet) \cite{wang_deep_2021} which runs multi-resolution convolutional layers connected in parallel---see \figref{hrnet} for a network diagram---and 3FabRec \cite{browatzki_3fabrec_2020}, which features a autoencoder ResNet structure that learns facial representations in an unsupervised manner before switching to supervised learning for facial landmark estimation. 

For HRNet, we performed extensive validation set experiments (not involving our final test data) to test a number of joint-training and fine-tuning paradigms, and a range of training data augmentations including random zooms and rotations, inspired by our earlier observations that landmark estimation is often poor for infants with difficult poses. Our tests yielded two final models with similar validation performance:

\noindent\textbf{HRNet-R90JT}, HRNet \textbf{j}ointly \textbf{t}rained on the union of the 300-W training set and InfAnFace Train, with \textbf{r}otations of angles between $\pm\textbf{90}^{\circ}$ applied to the training data 3/5ths of the time.

\noindent\textbf{HRNet-R90FT}, where HRNet is first trained on 300-W data, then \textbf{f}ine-\textbf{t}uned by further training on InfAnFace Train with a reduced learning rate and parameters frozen before the second HRNet layer, and with \textbf{r}otations of angles between $\pm\textbf{90}^{\circ}$ applied 3/5ths of the time for both phases of training.  

The hyperparameters and configurations that were chosen by validation included:
\begin{enumerate}
    \item the rotation augmentation range of $\pm90^{\circ}$, from the choices $\{\pm30^{\circ}, \pm 60^{\circ}, \pm 90^{\circ}, \pm 120^{\circ}, \pm 150^{\circ}\}$;
    \item the random zoom range of $100\% \pm 25\%$, from the choices $\{\pm 5\%, \pm 10\%, \pm 15\%, \pm 20\%, \pm 30\%, \pm 50\%\}$;
    \item for fine-tuning runs, the starting learning rate of $10^{-7}$, from the choices $\{10^{-4}, 10^{-5}, 10^{-6}, 10^{-7}, 10^{-8}\}$;
    \item for fine-tuning runs, the choice to freeze the model before the second HRNet layer, from choice of second, third, or fourth layers, or leaving the entire model unfrozen (chosen simultaneously with the previous parameter in an exhaustive grid search); and
    \item the choice of which epoch's checkpoint to take to represent the model, upon conclusion of a training run.
\end{enumerate} 

To demonstrate the necessity of the features employed in our best models above, we also trained three further HRNet models with some of these features selective ablated:

\noindent\textbf{HRNet-JT}, HRNet \textbf{j}ointly \textbf{t}rained on the union of the 300-W training set and InfAnFace Train, with HRNet's default rotations of angles between $\pm30^{\circ}$ applied to the training data 3/5ths of the time.

\noindent\textbf{HRNet-FT}, where HRNet is first trained on 300-W data, then \textbf{f}ine-\textbf{t}uned by further training on InfAnFace Train with a reduced learning rate and parameters frozen before the second HRNet layer, and with HRNet's default rotations of angles between $\pm30^{\circ}$ applied to the training data 3/5ths of the time.

\noindent\textbf{HRNet-R90}, HRNet trained only on the 300-W training set, with \textbf{r}otations of angles between $\pm\textbf{90}^{\circ}$ applied to the training data 3/5ths of the time.

Furthermore, to demonstrate the effectiveness of our technique more generally, we also applied these modifications to the 3FabRec model at the supervised training stage, obtaining:

\noindent\textbf{3FabRec-JT}, 3FabRec \textbf{j}ointly \textbf{t}rained on the union of the 300-W training set and InfAnFace Train.

\noindent\textbf{3FabRec-FT}, where 3FabRec is first trained on 300-W data, then \textbf{f}ine-\textbf{t}uned by further training on InfAnFace Train.

\noindent\textbf{3FabRec-R90}, 3FabRec trained only on the 300-W training set, with \textbf{r}otations of angles between $\pm\textbf{90}^{\circ}$ randomly applied.

\noindent\textbf{3FabRec-R90JT}, 3FabRec \textbf{j}ointly \textbf{t}rained on the union of the 300-W training set and InfAnFace Train, with \textbf{r}otations of angles between $\pm\textbf{90}^{\circ}$ randomly applied.

These configurations were chosen to mirror our HRNet configurations, rather than by validation hyperparameter search, to allow for direct comparisons.

\subsection{Experimental results and analysis}
The infant facial landmark estimation performance metrics for 3FabRec, HRNet, and the modified versions of those models are reported in \tabref{nme-table} and \tabref{fr-auc-table}, and sample landmark predictions for our best-performing HRNet-R90JT model can be found in \figref{landmark-predictions}. These results show significantly improved performance from our HRNet-R90JT and HRNet-R90FT models compared to the original 3FabRec and HRNet, bringing infant facial landmark estimation results within striking range of the best results on adult faces. The most notable gains come from performance on InfAnFace Challenge, but there are also small gains in performance on InfAnFace Common. A visualization of the the marked improvements of HRNet-R90JT over HRNet on a per-landmark level can be found in \figref{mean-face-errors}, and overall cumulative error curves can be found in Supp. \secref{ced-curves}. 

\figmeanfaceerrors

The reported results on the models with ablated features, HRNet-JT, HRNet-FT, and HRNet-R90, demonstrate that both the infant training data and the wider rotation angles are needed to realize the full gains. Intriguingly, though, either feature on its own seems capable of taming the most egregious predictions on InfAnFace Challenge. We also observe that the effect of applying wider rotation angles in training, on models trained on infant data, is to notably improve performance on InfAnFace Challenging faces at only a slight cost of performance on the largely-upright faces in InfAnFace Common.

The performance of the modified 3FabRec models are weaker than those of correspondingly modified modified HRNet models, perhaps in part due to the exhaustive hyperparameter validation tuning performed for the latter. However, the relative performances of the different 3FabRec variants support our general conclusions that while domain adaptation and data augmentation are both individually quite effective for improving performance, their combination leads to further gains; and that most gains come from improvements to the difficult InfAnFace Challenging images.

\section{Infant face detection}
\label{sec:face-detection}

Facial landmark estimation algorithms usually require information about face location as input, typically in the form of coordinates for a bounding box. These coordinates are included with test datasets like InfAnFace Test, but for most real-world applications, they need to be obtained beforehand from dedicated \emph{face detection} algorithms. While in principle, such algorithms could be fine-tuned with infant data, we found them to be already adequate out-of-the-box.

To demonstrate this, we applied a cutting-edge face detection algorithm, RetinaFace \cite{deng_retinaface_2019}, to InfAnFace Test. For face detection, the main performance metric on a set of predictions is the \textit{average precision}, defined as an interpolated area under an associated precision-recall curve (see Supp. \secref{average-precision}). The average precisions of RetinaFace across the InfAnFace Test, Common, and Challenging sets respectively are 98.1\%, 100.0\%, and 94.7\%, attesting to excellent performance on infants. Thus, the combination of RetinaFace and our HRNet-R90FT and HRNet-R90JT models provide complete solutions to the facial landmark estimation task for infants.

\section{An in-field baby monitor challenge}
\label{sec:baby-monitor}

As part of an ongoing study exploring links between infant motor function and developmental disruptions, our lab gathered baby monitor footage of 15 real infant subjects under an Institutional Review Board (IRB \#17-08-19) approval. The videos feature infants napping and sleeping in their own crib or bassinet, typically under low-light conditions, triggering the baby monitor's monochromatic infrared capture mode. These pose and lighting conditions entail significant challenges to algorithmic comprehension. 

We annotated 213 private images from these videos in order to gauge facial landmark estimation performance in extremely adverse vision conditions. The predictions of the original HRNet and of our improved HRNet-R90JT on this set yielded respective NMEs of 32.3 and 15.5 under the minimal box size normalization\footnote{Note: The interocular norm is not suitable in this setting because many in-bed infants have their heads turned, hiding one eye, and the industry convention of assigning that eye's landmark to the nearest visible spot on the head renders the interocular distance somewhat arbitrary.}, poor in comparison to HRNet-R90JT's error of 3.73 on the InfAnFace Challenging set.

We attempted to mitigate this performance gap with further data augmentation techniques. We established a new model, \textbf{HRNet-R150GJT}, \textbf{j}ointly \textbf{t}rained on 300-W and InfAnFace Train data, with \textbf{r}otation augmentations in the range of $\pm \textbf{150}^{\circ}$, and a \textbf{g}rayscale filter applied 1/2 of the time, where both hyperparameters were chosen based on validation on InfAnFace Train. The resulting predictions yielded an average NME of 11.7 under the box normalization factor. This is a notable improvement, but deeper developments are needed to raise performance to a higher level.

\section{Conclusion and future work}

We have advanced the algorithmic comprehension of infant faces by introducing the first dataset in the field, demonstrating the inadequacies of existing facial landmark estimation algorithms on the infant domain, and training best-in-class models using domain adaptation techniques and our newly introduced training data. We also demonstrated the effectiveness of existing face detection algorithms on infants, yielding a full-stack solution to the landmark estimation problem for infants. Finally, we explored applications to challenging in-field data, and pointed out the need for further research in this area.

\bibliographystyle{IEEEbib}
\bibliography{paper}

\supp
\end{document}